\documentclass[a4paper,fleqn]{sc}

\usepackage{graphicx}
\usepackage{caption}
\usepackage{subcaption}
 \usepackage{tikz}
\usetikzlibrary{positioning, shapes.geometric, arrows, arrows.meta}
\usepackage[authoryear]{natbib}
\def\tsc#1{\csdef{#1}{\textsc{\lowercase{#1}}\xspace}}
\tsc{WGM}
\tsc{QE}
\tsc{EP}
\tsc{PMS}
\tsc{BEC}
\tsc{DE}

\begin{document}
\let\WriteBookmarks\relax
\def\floatpagepagefraction{1}
\def\textpagefraction{.001}
\shorttitle{Support Vector Regression Parameters Optimization using Golden Sine Algorithm and its application in stock market}
\shortauthors{Mohammadreza Ghanbari, Mahdi Goldani}

\title [mode = title]{Support Vector Regression Parameters Optimization using Golden Sine Algorithm and its application in stock market}                      

\author[1]{\color{black}Mohammadreza Ghanbari}
\ead{ghi.mohammadr@gmail.com}



\address[1]{Department of Mathematical Sciences , Sharif University of Technology, Tehran, Iran}

\author[2]{\color{black}Mahdi Goldani}
\cormark[1]
\ead{m.goldani@hsu.ac.ir}


\address[2]{Department of Economics, Hakim Sabzevari University, Sabzevar, Iran}



\begin{abstract}
Support vector machine modeling is a new approach in machine learning for classification showing good performance on forecasting problems of small samples and high dimensions. Later, it promoted to Support Vector Regression (SVR) for regression problems. A big challenge for achieving reliable is the choise of appropriate parameters. Here, a novel Golden sine algorithm (GSA) based SVR is proposed for proper selection of the parameters. For comparison, the performance of the proposed algorithm is compared with eleven other meta-heuristic algorithms on some historical stock prices of technological companies from Yahoo Finance website based on Mean Squared Error and Mean Absolute Percent Error. The results demonstrate that the given algorithm is efficient for tuning the parameters and is indeed competitive in terms of accuracy and computing time.

\end{abstract}



\maketitle

\section{Introduction}
Support Vector Machine (SVM) was originally introduced by Vapnik \citep{cortes1995support} for classification problems. It can potentially solve small-sample, non-linear and high dimensions problems by using structural risk minimization principle instead of the empirical risk principle. Essentially, the SVM is a convex quadratic programming method by which it is possible to find the global,  rather than local optimum. After on, SVM was developed and extended to Support Vector Regression (SVR) to solve regression problems \citep{cortes1995support}. SVR is arguably one of the best techniques and experimental results show considerable performance compared to other nonlinear methods~\citep{ thissen2003using, vapnik2013nature}. However, setting of the parameters for the SVR plays a significant role and the performance accuracy changes considerably upon a bad choice of parameters~\citep{chapelle2002choosing, duan2003evaluation, kwok2000evidence,  yeh2011multiple}. 

Concerning this issue, a main approach to select the SVR parameters optimally is to make use of an optimization technique for finding  optimal values. Generally, three common techniques to optimize the SVR parameters are grid search~\citep{hsu2003practical}, which in practical applications, it is usually vulnerable to get to a local optimum, gradient descent~\citep{keerthi2007efficient} and meta-heuristics algorithms~\citep{blum2003metaheuristics, talbi2009metaheuristics}.

Meta-heuristics algorithms have been introduced that may provide a sufficiently good solution to an optimization problem specially on presence of incomplete information or limited computing capacity. They have shown superior results in the case of solving optimization problems for parameter tuning of complex models  ~\citep{gogna2013metaheuristics}. 
In the past, many meta-heuristics 
algorithms have been proposed for the selection of optimal SVR parameters. Example are, Genetic Algorithm (GA)~\citep{huang2012hybrid, gu2011housing, min2006hybrid, wu2009novel}, Grey Wolf Optimizer (GWO)~\citep{mustaffa2015ls}, Particle Swarm Optimization (PSO)~\citep{li2011tax, wu2010hybrid},
Sine Cosine Algorithm (SCA)~\citep{li2018parameter},
Butterfly Optimization Algorithm (BOA) \citep{ghanbari2019forecasting},
Firefly Algorithm (FA)~\citep{kavousi2014new} and
Bat Algorithm (BA)~\citep{tavakkoli2015novel}.


Recently, a novel math-based meta-heuristic optimization algorithm inspired by sine function, named as Golden Sine Algorithm (GSA), was designed by \cite{tanyildizi2017golden}. The GSA algorithm searches to approach a better solution in each iteration by trying to bring the current point closer to the target value and the solution space gets to be narrowed by the golden section algorithm so that the areas with supposedly good results instead of the whole solution space are examined.

Here, we propose a novel GSA based SVR model where GSA is used to set the parameters of SVR. For validation, eleven other meta-heuristic algorithms, namely Whale Optimization Algorithm (WOA), Salp Swarm Algorithm (SSA), Neural Network Algorithm (NNA), Firefly Algorithm (FA), Multi-Verse Optimizer (MVO), Moth-Flame Optimization (MFO), Harris Hawks Optimization (HHO), Grey Wolf Optimization (GWO), Butterfly Optimization Algorithm (BOA), Biogeography-Based Optimization (BBO) and Artificial Bee Colony Optimization (ABC) are appropriated to SVR to optimize the parameters, are compared with the proposed algorithm.

The remainder of our work is as follows: In Section \ref{method}, we discuss the presented GSA-SVR model, the support vector regression and GSA. In Section \ref{experiment}, the  model is tested on some datasets and compared 
with other models and the obtained experimental results are discussed. Conclusions and future research directions are provided in Section \ref{con}.





\section{Materials and Methods}\label{method}
In this section, we briefly discuss about Support Vector Regression (SVR) and the Golden Sine Algorithm (GSA).


\subsection{Support Vector Regression}

Support vector machine (SVM) is a machine learning algorithm introduced by Vapnik in 1995 for classification problems. It has been one of the more widely used methods in recent years as a powerful method. It was first used to address a binary pattern classification problem. Then, it was promoted to support vector regression (SVR) for regression problems by using $\epsilon$-insensitive loss function to penalize data when they are greater than $\epsilon$~\citep{cortes1995support}. SVR aims to provide a nonlinear mapping function to map the training dataset to a high dimensional feature space \citep{yeh2011multiple}.

The given training dataset is
$\{(x_i,y_i)\}_{i = 1}^n$, where 
$x_i \in \mathbb{R}^d$
is input data,
$y_i \in \mathbb{R}$ is the output value of the $i$-th data point in the dataset, $d$ is the dimension of samples and $n$ is the number of samples. The nonlinear function between
the input and the output is formulated as:
\begin{align} \label{2.1}
y=f(x)=w^{T} \phi(x)+b,
\end{align}
where $\phi: \mathbb{R}^d \longrightarrow F$ is a nonlinear mapping to the feature space, $w \in F$ is a vector of weight coefficients and $b$ is a bias constant. The $w$ and $b$ are estimated by minimizing the following optimization problem:
\begin{equation}
\begin{split}
\text{Min} & \frac{1}{2} ||w||^2, \\
\text{s.t. } &
\begin{cases}
y_i-w^T\phi(x_i)-b  & \leq \epsilon,\\ 
y_i-w^T\phi(x_i)-b  & \geq -\epsilon,
\end{cases}
\end{split}
\end{equation}

The slack variables $\xi$ and $\xi^*$ are used to penalize points from $\epsilon$-insensitive band:
\begin{equation}
\begin{split}\label{2.3}
\text{Min} & \frac{1}{2} ||w||^2+ C\sum_{i=1}^n(\xi_i+\xi_i^*),\\
\text{s.t. } &
\begin{cases}
y_i-w^T\phi(x_i)-b  \leq \epsilon + \xi_i,\\ 
y_i-w^T\phi(x_i)-b  \geq -\epsilon - \xi_i^*,\\
\xi_i,\xi_i^* \geq 0, i = 1,\cdots, n,
\end{cases}
\end{split}
\end{equation}
where $C$ is a constant known as the penalty parameter to specify the
trade-off between the empirical risk and regularization terms, $\epsilon$ is the insensitive loss function and the slack variables $\xi_i$ and $\xi_i^*$, correspond to upper and lower deviations, respectively, and $n$ is the number of training
patterns.

Using the Lagrangian and corresponding optimality conditions, the obtained generic equation is written as \citep{smola2004tutorial, cortes1995support}:
\begin{align}
f(x)=\sum_{i=1}^{n}(\beta_i-\beta_i^*)K(x_i,x)+b,
\end{align}
where $\beta_i$ and $\beta_i^*$ are nonzero Lagrange multipliers and
$K(x_i,x)$ is the kernel function. In our work, Radial Basis Function (RBF) has been used as kernel function:
\begin{align}
K(x_i,x_j) = \exp(-\gamma||x_i-x_j||^2),
\end{align}
where $\gamma$ is the RBF width parameter.

\subsection{Golden Sine Algorithm}
Golden Sine Algorithm (GSA) is a novel math-based meta-heuristic optimization algorithm inspired by sine function for solving optimization problems \citep{tanyildizi2017golden}.
Sine, a trigonometric function, is the coordinate relative to the $y$-axis of a point on a $1$-unit radius circle that is the central origin. An orthogonal triangle with an angle made by the $y$-axis of a straight line drawn from the origin or with the same angle is calculated with the hypotenuse section of the edge opposite this angle. The defining range of the function is $[-1, 1]$. 
The scan of the unit circle of all values of the sine
function is similar to the search of the search space in
optimization problems. This similarity has inspired the
development of GSA. The operator used in the algorithm is
shown by

\begin{equation}\label{eq: 2.3.1}
V_{ij}=V_{ij}\left|\sin \left(r_{1}\right)\right|-r_{2} \sin \left(r_{1}\right)\left|x_{1} D_j-x_{2} V_{ij}\right|
\end{equation}
where $V_{ij}$ is the value of current solution in the $i$-th dimension, $D$ is the determined target value, $r_1$ is a random number in the range $[0, 2\pi]$ and $r_2$ is a random number in the range $[0, \pi]$, and $x_1$ and $x_2$ are the coefficients obtained by the golden section method. These coefficients limit the search space and also allow the current value to approach the target value. 

The wide range of the search space is a major problem
for solving problems. The effect of limiting the search
space in solving problems is significantly affecting the
results. GSA uses the golden section method to make
this process the best possible way. Golden section search is an optimization technique that can be used to find the maximum or minimum value of a single unimodal function. The name is from the golden ratio. Two numbers, $p$ and $q$, are in a golden ratio if 

\begin{equation}
\frac{p+q}{p}=\frac{p}{q}=\tau,
\end{equation}

or equivalently,

\begin{equation}
1+\frac{q}{p}=\tau,
\end{equation}

or

\begin{equation}\label{eq: 2.3.5}
1+\frac{1}{\tau}=\tau
\end{equation}

Solving Eq. \ref{eq: 2.3.5}, we get the positive root as

\begin{equation}\label{eq: 2.3.6}
\tau=\frac{1-\sqrt{5}}{2} \approx 0.618033,
\end{equation}
here $\tau$ is called the golden number.

In GSA, initial default values for $a$ and $b$ are
considered to be $-\pi$ and $\pi$, respectively. These two
coefficients are applied to the current and target values in
the first iteration. Then, the coefficients $x_1$ and $x_2$ are
updated as the target value changes.

To avoid the situation of equality for $x_1$ and $x_2$, an equality check is performed. It means that if the two values are equal, then the random numbers $rand_1$ and  $rand_2$ are generated in the range, respectively, $[0, \pi]$ and $[0, -\pi]$ and $x_1$ and $x_2$ are recalculated.

\subsection{GSA for Parameter Optimization of SVR}

Before presentation the algorithm, we first discuss the data pre-processing of time series.

Phase Space Reconstruction \citep{takens1981detecting} is a method in which uncover the hidden information embedded in the time series dynamics. This method provides a simplified multidimensional representation of data.

Let $ \{x_i\}_{i=1}^n$ represent an $n$ point time series.
Then, the reconstructed phase space can be expressed as a matrix as follows:
\begin{equation}
X =
\begin{bmatrix}\label{xtakens}
x_1 & x_{1+\tau} & \cdots &  x_{1+(m-1)\tau}\\
x_2 & x_{2+\tau} & \cdots & x_{2+(m-1)\tau}\\
 \vdots & \vdots & \ddots & \vdots \\
x_{n-1-(m-1)\tau} & x_{n-1-(m-2)\tau} & \cdots & x_{n-1}
\end{bmatrix},
\end{equation}
where $\tau$ is the time delay constant and $m$ is called the embedding dimension of the reconstructed phase space. 

\citep{kennel1992determining} proposed an efficient method of finding the minimal sufficient embedding dimension, named as false nearest neighbors (FNN) procedure, in which the nearest neighbors of every point in a given dimension are found, and then checks are made to see if these points are still close neighbors in one higher dimension. To estimate the delay parameter, here we use the first minimum of the Mutual Information (MI) function \citep{abarbanel2012analysis}.

After finding the optimal $m$ and $\tau$, the input data and the output vector were designed by Eq. (\ref{xtakens}) and Eq. (\ref{ytakens}).
\begin{equation}\label{ytakens}
Y =
\begin{bmatrix} 
Y_1\\
Y_2\\
\vdots\\
Y_n 
\end{bmatrix}
=
\begin{bmatrix}
x_{2+(m-1)\tau}\\
x_{3+(m-1)\tau}\\
\vdots\\
x_n
\end{bmatrix} 
\end{equation}

After construction of the input and output matrix, we normalize the data by using min-max formula
\begin{equation}\label{2.4}
x_{\text{new}} = \frac{x_{\text{old}}-x_{\text{min}}}{x_{\text{max}}-x_{\text{min}}},
\end{equation}
to scale to the range $[0,1]$. Finally, the data is divided into the training set and the testing set.

In the first step of our proposed algorithm, the GSA parameters including the number of agents and maximum number of iterations are set. Then, GSA-SVR starts with a set of candidate solutions generated randomly within predetermined lower and upper bounds. In this case, each solution is a three-dimensional vector represented by ($C$, $\gamma$, $\epsilon$), where $C$, $\gamma$ and $\epsilon$ are the SVR parameters to be optimized. The objective function is equal to the Mean Square Error (MSE), Eq. \ref{2.6}, of the tested SVR model. A predetermined maximum number of iterations is used as a criteria to stop the algorithm. Figure \ref{flochart} shows flowchart of the complete the complete procedure.

A step-wise procedure of the proposed algorithm is described next:

\begin{itemize}[leftmargin=25mm]
\item[\textbf{Step 1:}]
Assign the parameters including the number of search agents and the maximum number of iterations. Set the iteration number, $t$, equal to zero.
\item[\textbf{Step 2:}]
Initialize the random solutions of search agents with  
\begin{equation}\label{2.5}
S_i = Lb_i + (Ub_i-Lb_i)\cdot u
\end{equation}
and evaluate fitness function using Eq. (\ref{2.6}) and Eq. (\ref{2.7}) on the test data. Here, $Lb$ $(Ub)$ is lower (upper) bound and $i \in \{C, \gamma, \epsilon\}$, and $u$ is a uniform random number in the interval $(0,1)$.
\item[\textbf{Step 3:}]
Update the position of search agents for every dimension based on Eq. (\ref{eq: 2.3.1}) and set $t=t+1$.
\item[\textbf{Step 4:}]
If the maximum number of iterations is reached then the optimized parameters of SVR are selected, thus go to step 5; otherwise go back to step 3.
\item[\textbf{Step 5:}]
Use the SVR model with the optimal parameters ($C$, $\gamma$, $\epsilon$) for prediction.


\end{itemize}

\tikzstyle{rect} = [draw = black, rectangle, fill=white!20, text width=10em, text centered, minimum height=2em, text= black]
\tikzstyle{elli} = [draw, ellipse, fill=white!20, minimum height=2em, text= black]
\tikzstyle{circ} = [draw, circle, fill=white!20, minimum width=8em, inner sep=10pt, text= black]
\tikzstyle{diam} = [draw = black, diamond, fill=white!20,  minimum width=6em,  minimum height=2em, text width=6em, text height=0.05em, text badly centered, inner sep = 0pt, text= black]
\tikzstyle{line} = [draw, -latex', text= black]

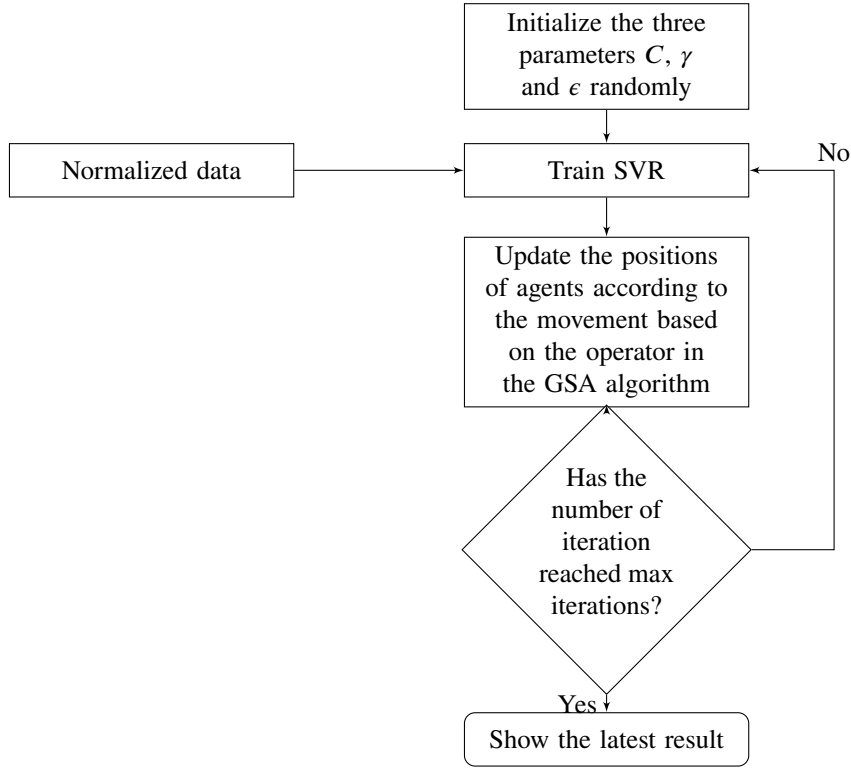
\begin{figure}[h]
\begin{center}
\begin{tikzpicture}[node distance = 1.5cm]
\node  [rect](step2){Initialize the three parameters $C$, $\gamma$ and $\epsilon$ randomly};
\node  [rect, below of = step2] (step3) {Train SVR};
\node  [rect, left of = step3, node distance=6cm] (step4) {Normalized data};
\node  [rect, below of = step3, node distance=2cm] 
(step5) {Update the positions of agents according to the movement based on the operator in the GSA algorithm};
\node  [diam, below of = step5, node distance=3cm] (step6) {Has the number of iteration reached max iterations?};
\node  [rect, rounded corners, below of = step6, node distance=2.5cm] (step7) {Show the latest result};
\path [line, color=black] (step2) -- (step3);
\path [line, color=black] (step3) -- (step5);
\path [line, color=black] (step4) -- (step3);
\path [line, color=black] (step5) -- (step6);
\path [line, color=black] (step6) -| +(3,0) |- node [anchor=south] {No} (step3);
\path [line, color=black] (step6) -- node [anchor = east] {Yes} (step7);
\end{tikzpicture}
\caption{GSA-SVR procedure}
\label{flochart}
\end{center}
\end{figure}

\section{Experimental Results}\label{experiment}

In this section, a number of stocks are chosen to test the performance of our proposed GSA-SVR model. The proposed algorithm is compared to the other meta-heuristic algorithms, being used for parameter optimization of SVR, including Whale Optimization Algorithm based SVR (WOA-SVR), Salp Swarm Algorithm based SVR (SSA-SVR), Neural Network Algorithm based SVR (NNA-SVR), Firefly Algorithm based SVR (FA-SVR), Multi-Verse Optimizer based SVR (MVO-SVR), Moth-Flame Optimizer based SVR (MFO-SVR), Harris Hawks Optimization Algorithm based SVR (HHO-SVR), Grey Wolf Optimization Algorithm based SVR (GWO-SVR), Butterfly Optimization Algorithm based SVR (BOA-SVR), Biogeography-Based Optimization Algorithm based SVR (BBO-SVR) and Artificial Bee Colony Optimization Algorithm based SVR (ABC-SVR).

Stock market price prediction is regarded as one of the most challenging tasks of financial time series prediction. The difficulty of forecasting arises from the inherent non-linearity and nonstationarity of the stock market and financial time series.
Thus, daily closing stock market prices of three companies, namely Alibaba Group Holding Limited (BABA) (from 03/10/2016 to 01/10/2019), Tesla, Inc. (TSLA) (from 03/10/2016 to 01/10/2019) and Taiwan Semiconductor Manufacturing (TSM) Company Limited (from 03/10/2016 to 01/10/2019), were extracted from Yahoo Finance historical quotes. After finding the time delay, $\tau$, the embedding dimension, $m$ and reconstructing the phase space, $80\%$ of the data were used as the training set and the remaining were used as the testing set. All the predictions were based on one-step ahead prediction results and the computations were carried out in MATLAB R2019a environment using the LIBSVM Toolbox \citep{chang2011libsvm} on a laptop with an Intel(R) Core(TM) i3-3110M CPU @ 2.40GHz and 4 Gbytes memory.

In our work, Mean Squared Error (MSE) and Mean Absolute Percent Error (MAPE) were used in order to calculate the accuracy,
\begin{align}
    \text{MSE} &= \frac{1}{N}\sum_{i=1}^{N}(y_i - f_i), \label{2.6}\\
    \text{MAPE} &= \frac{1}{N}\sum_{i=1}^{N}\big|\frac{y_i - f_i}{y_i} \big|, \label{2.7}
\end{align}
where $y_i$ and $f_i$ denote the actual and predicted values for the $i$-th data point, respectively and $N$ is the number of forecasting days.

Since meta-heuristics algorithms use initial random population, we ran each algorithm several times to get the optimal answer. However, to increase the probability of finding the global optimum we used diversity in population and a sufficiently large number of iterations. In this study, the number of populations and the maximum number of iterations are selected to be 20 and 50 respectively. Also, the search space for both parameters $C$ and $\gamma$ were $[4^{-7},4^4]$ and the range for parameter $\epsilon$ was $[4^{-7}, 0.25]$. 

All details of datasets including name, embedding dimension $m$ and time delay $\tau$  are shown in Table \ref{table 1}. The size of training datasets is equal to 530 and the size of testing datasets is equal to 133 for all of three datasets. For phase space reconstruction we used the recurrence plot and recurrence quantification analysis of MATLAB toolbox \citep{chen2012multiscale}.

\begin{table}[width=1.0\linewidth, cols=6, pos=h]
  \centering
  \caption{Estimation of $m$ and $\tau$ for phase space reconstruction.}
    \begin{tabular*}{\tblwidth}{@{}CCCCCC@{} }
    \toprule
      \textbf{Parameters}     & \textbf{ BABA } &       & \textbf{ TSLA } &       & \textbf{ TSM} \\
    \midrule
    $m$ &  10    &       &  12    &       &  10 \\
    $\tau$ &  10     &       &  2     &       &  5 \\
    \bottomrule
    \end{tabular*}%
  \label{table 1}%
\end{table}%

\begin{figure}[h]
\centering
\hspace*{0.2in}
\begin{subfigure}{.4\textwidth}
  \includegraphics[width=1.3\linewidth]{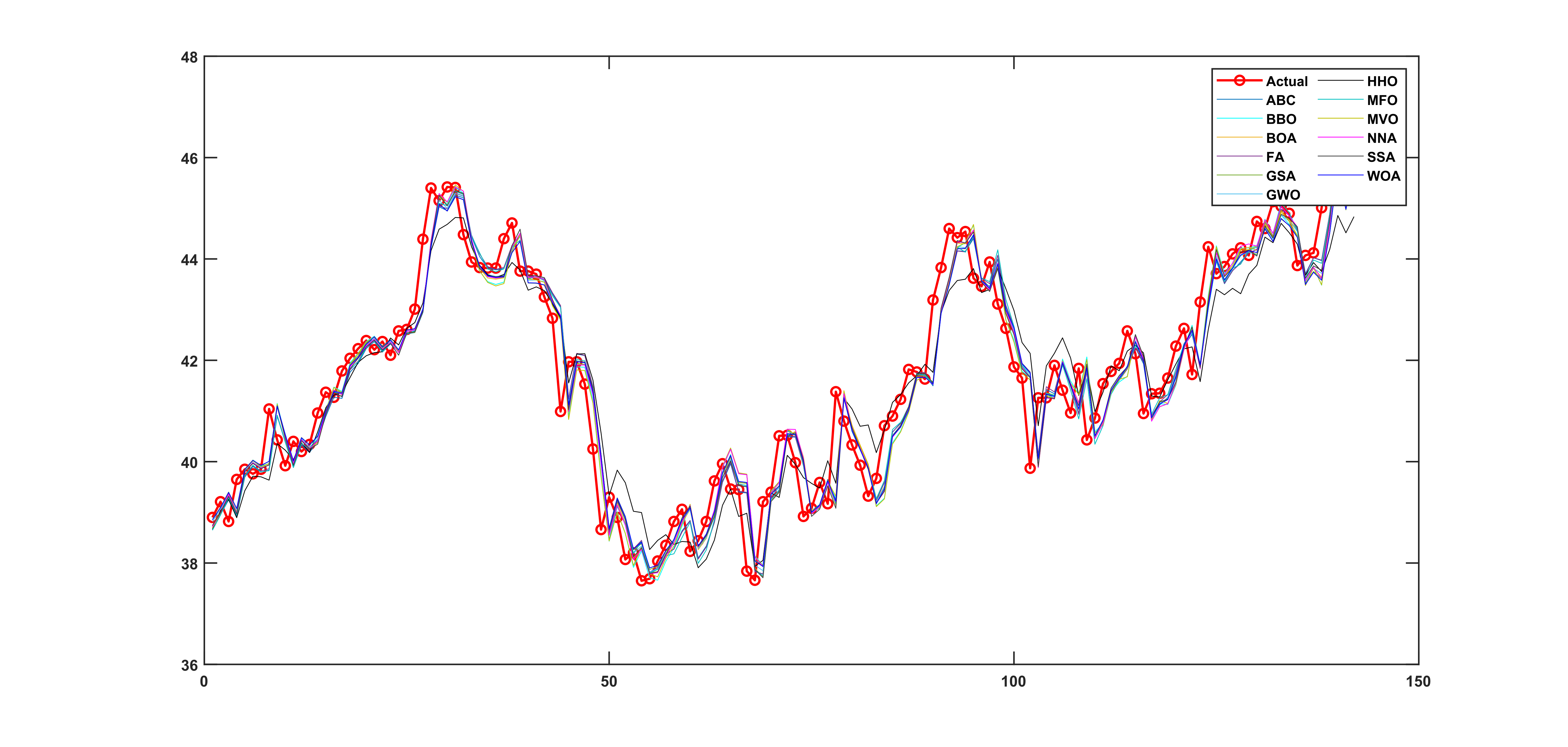}
  \caption{TSM}
  \label{tsmplot}
\end{subfigure}
\newline
\centering
\begin{subfigure}{.4\textwidth}
\hspace*{-0.9in}
  \includegraphics[width=1.3\linewidth]{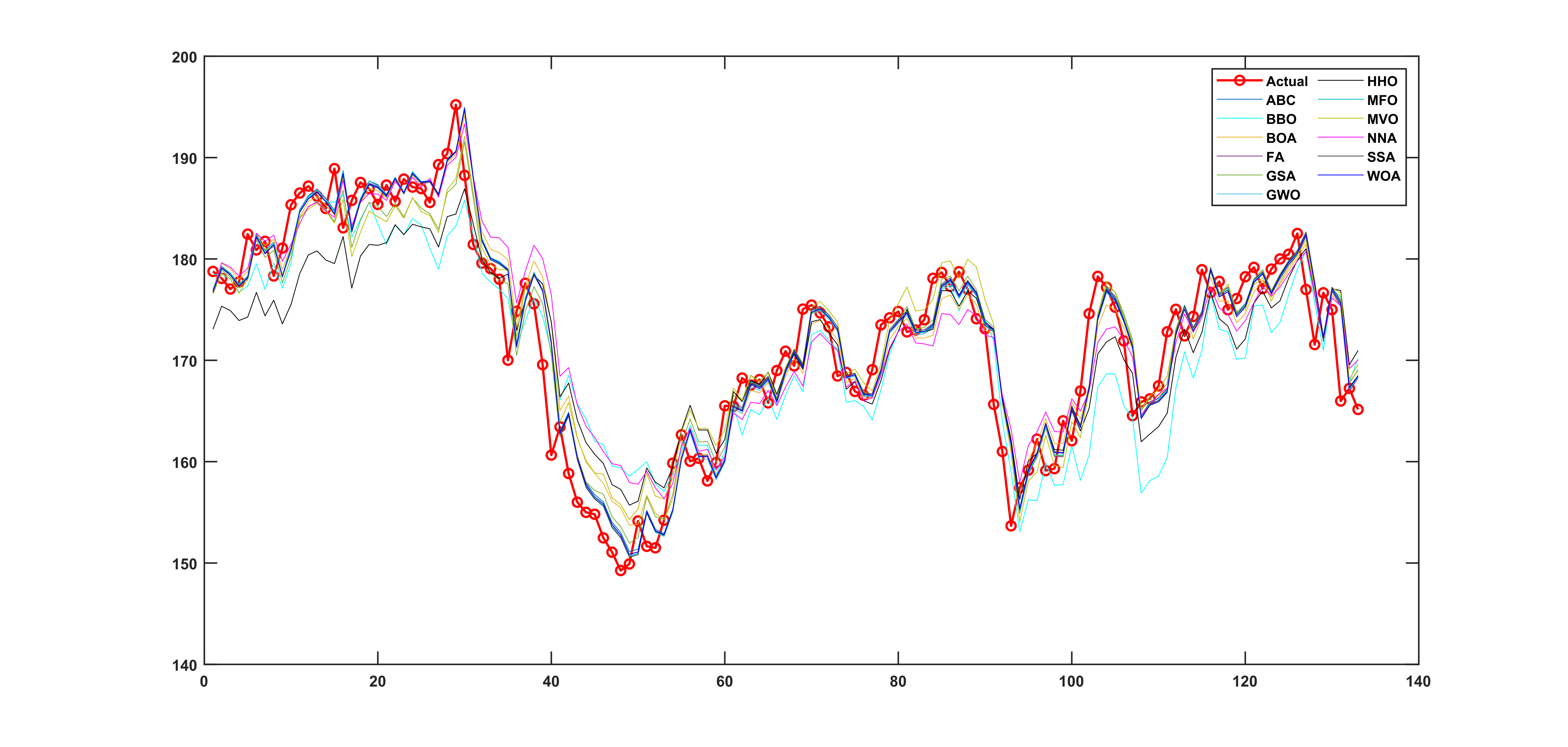}
  \centering \caption{BABA}
    \label{babaplot}
\end{subfigure}
\centering
\begin{subfigure}{.4\textwidth}
\hspace*{-0.1in}
  \includegraphics[width=1.3\linewidth]{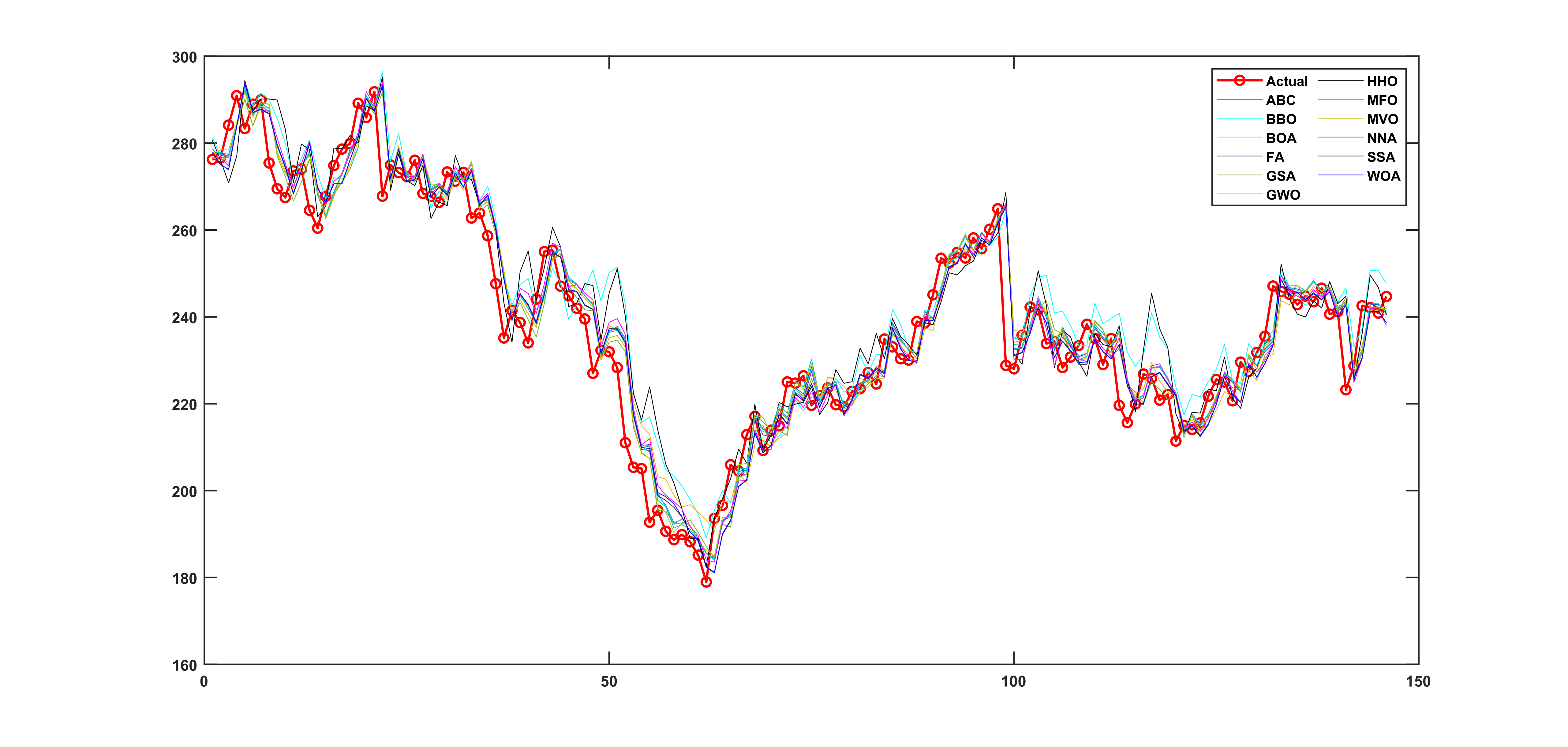}
  \centering \caption{TSLA}
     \label{tslaplot}
\end{subfigure}
\caption{Prediction comparison for datasets a) TSM b) BABA and c) TSLA show that among the twelve algorithms, GSA-SVR performs as one of the best based on accuracy. }
\label{allTests}
\end{figure}

Actual and predicted values obttained by our model compared to eleven other methods for the three datasets TSM (Fig. (\ref{tsmplot})), BABA (Fig. (\ref{babaplot})) and TSLA (Fig. (\ref{tslaplot})) are illustrated in Fig. \ref{allTests} after de-normalization. Also, Table \ref{babatable} presents the optimal values for the three parameters $C, \gamma$ and $\epsilon$, as well as MSE, MAPE and the computing time for BABA testing dataset for all of SVR-based methods. The same results for TSLA and TSM testing datasets are shown in Tables \ref{tslatable} and \ref{tsmtable}, respectively.

\begin{figure}[h]
\centering
\hspace*{0.2in}
\begin{subfigure}{.4\textwidth}
  \includegraphics[width=1.3\linewidth]{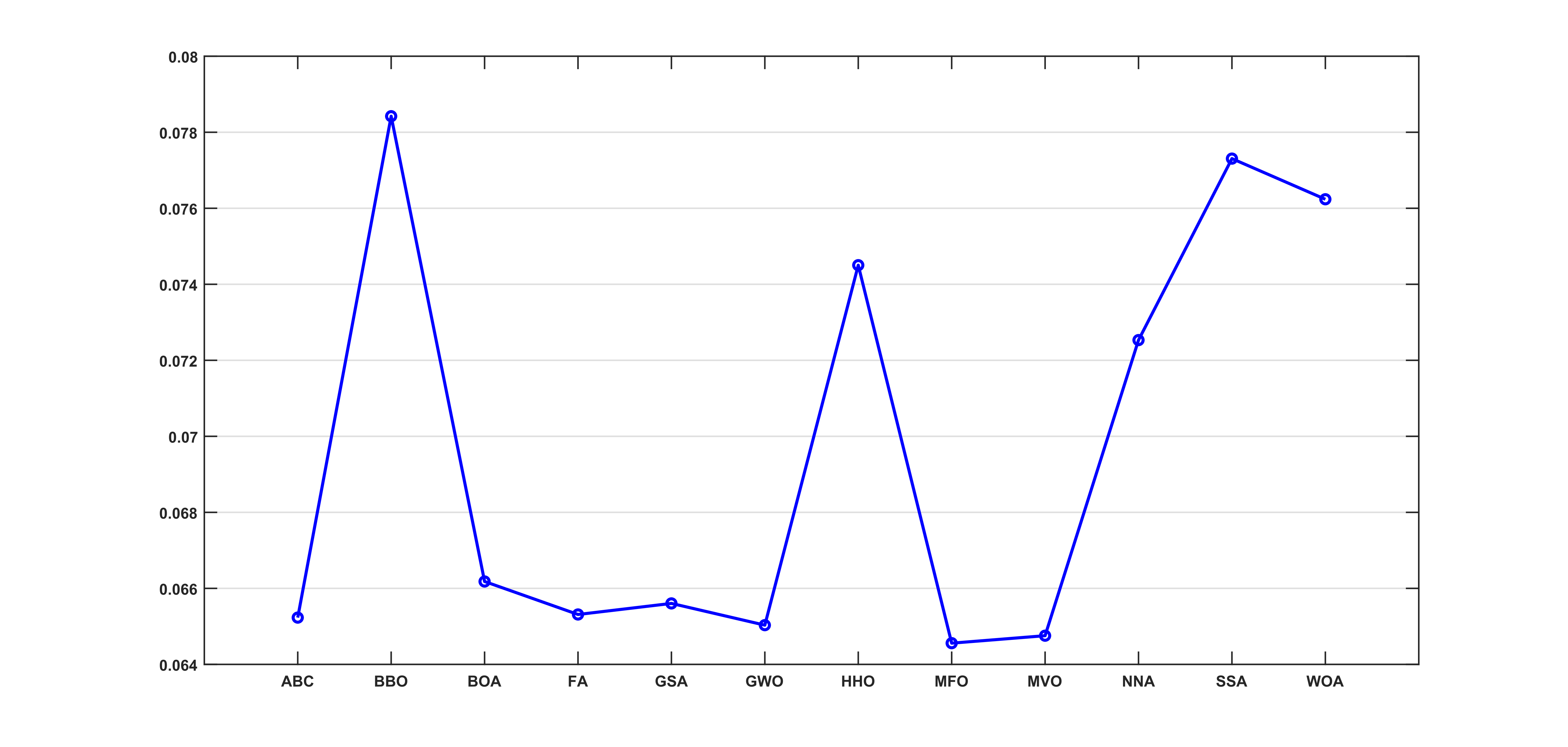}
  \caption{MAPEs}
  \label{mape}
\end{subfigure}
\newline
\centering
\begin{subfigure}{.4\textwidth}
\hspace*{-0.9in}
  \includegraphics[width=1.3\linewidth, height=4.5cm]{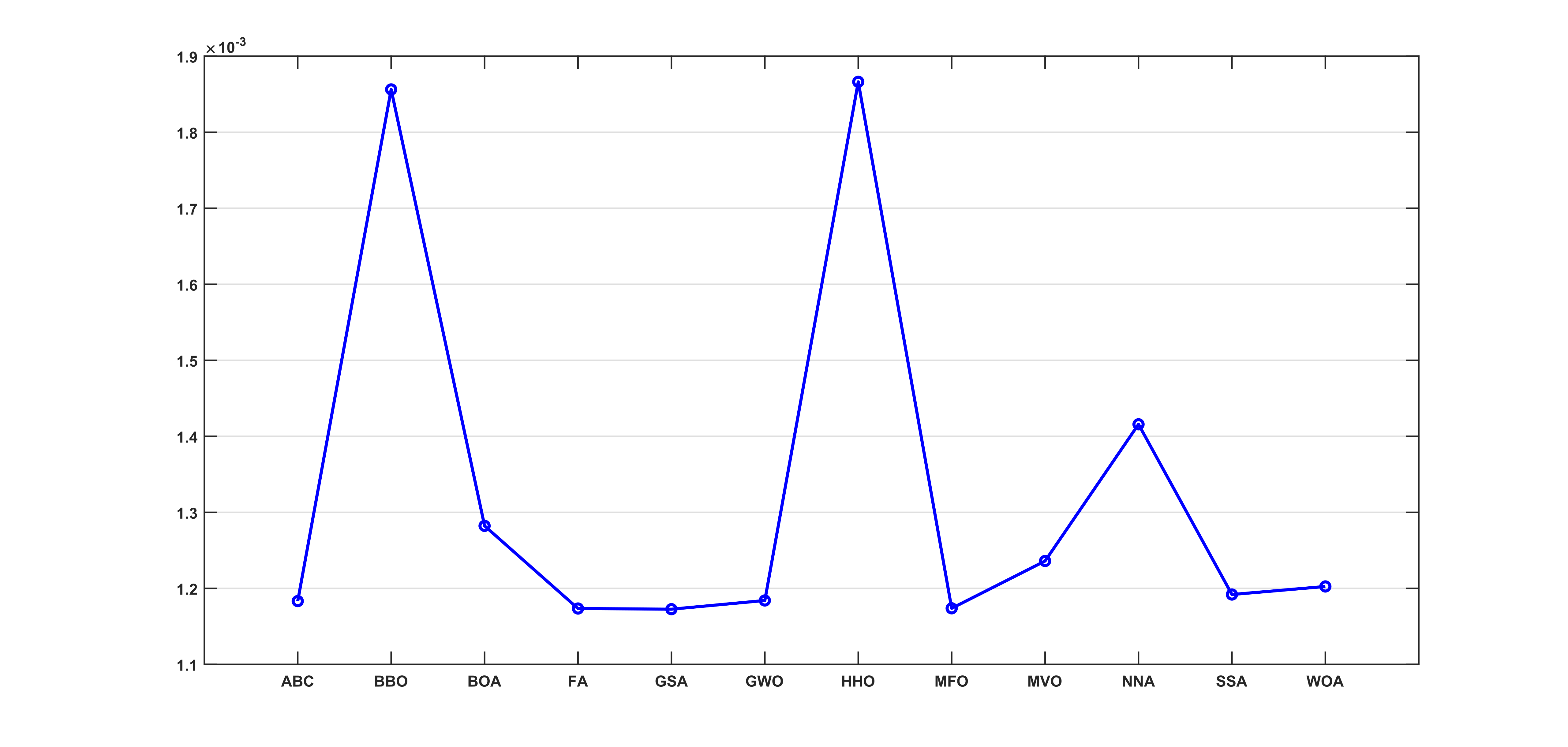}
  \centering  \caption{MSEs}
    \label{mse}
\end{subfigure}
\centering
\begin{subfigure}{.4\textwidth}
\hspace*{-0.1in}
  \includegraphics[width=1.3\linewidth, height=4.5cm]{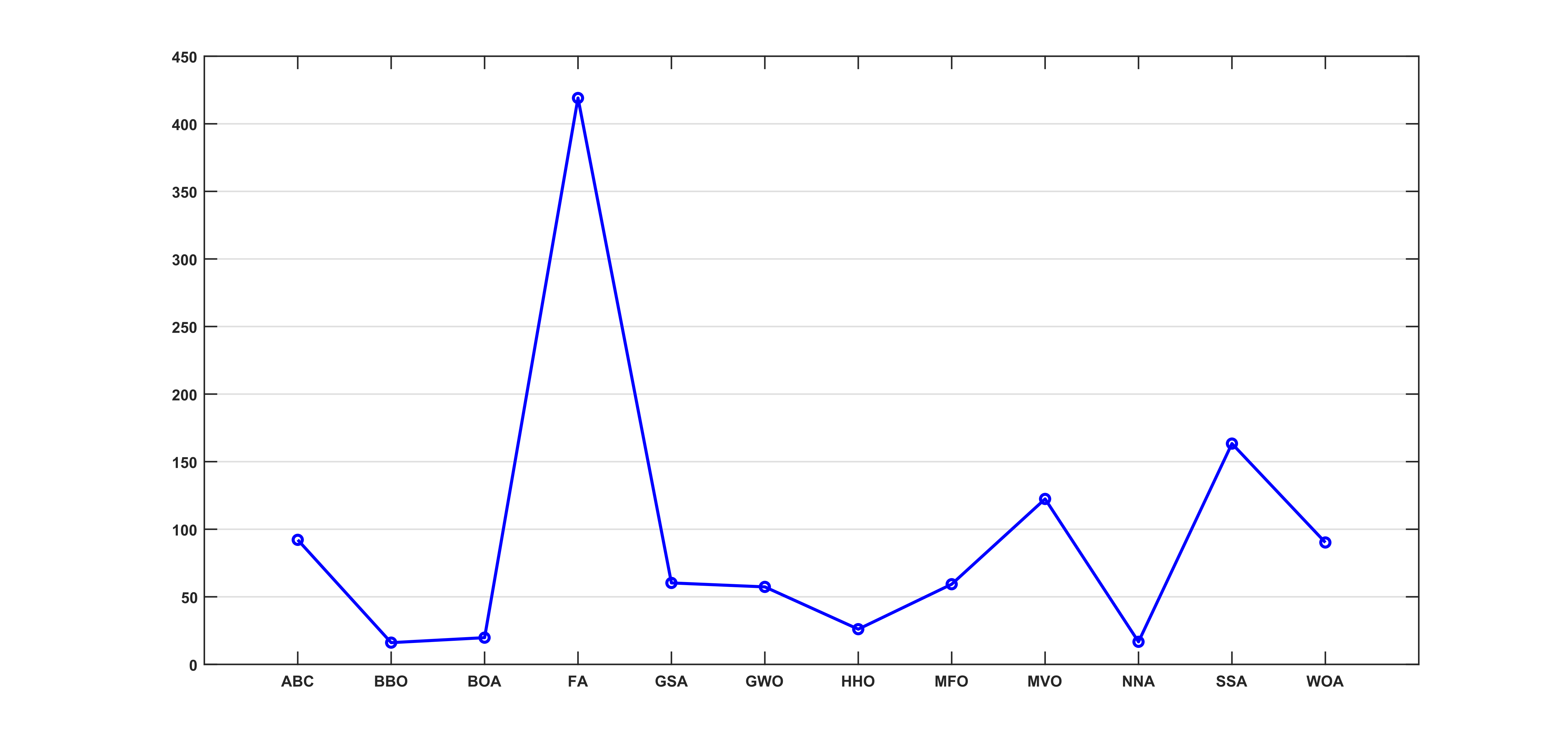}
  \centering \caption{Time Consumption}
     \label{timecosts}
\end{subfigure}
\caption{Fig. \ref{mape} and Fig. \ref{mse} compare the MAPE and the MSE of the twelve methods. Also Fig. \ref{timecosts} compares the cost time.}
\label{errorsmean}
\end{figure}

We now discuss results of our proposed algorithm in comparison with others. As shown in Fig. \ref{mape}, the MAPE of GSA-SVR ranked 6th with a slight difference of 0.001 with the first rank, MFO-SVR. Even though MAPE accuracy of GSA-SVR is slightly below the other five methods MFO-SVR, MVO-SVR, GWO-SVR, ABC-SVR and FA-SVR, its computing time is significantly better than these methods except for GWO-SVR.

\begin{figure}[h]
\centering
\hspace*{0.2in}
\begin{subfigure}{.4\textwidth}
  \includegraphics[width=1.3\linewidth,height=4.5cm]{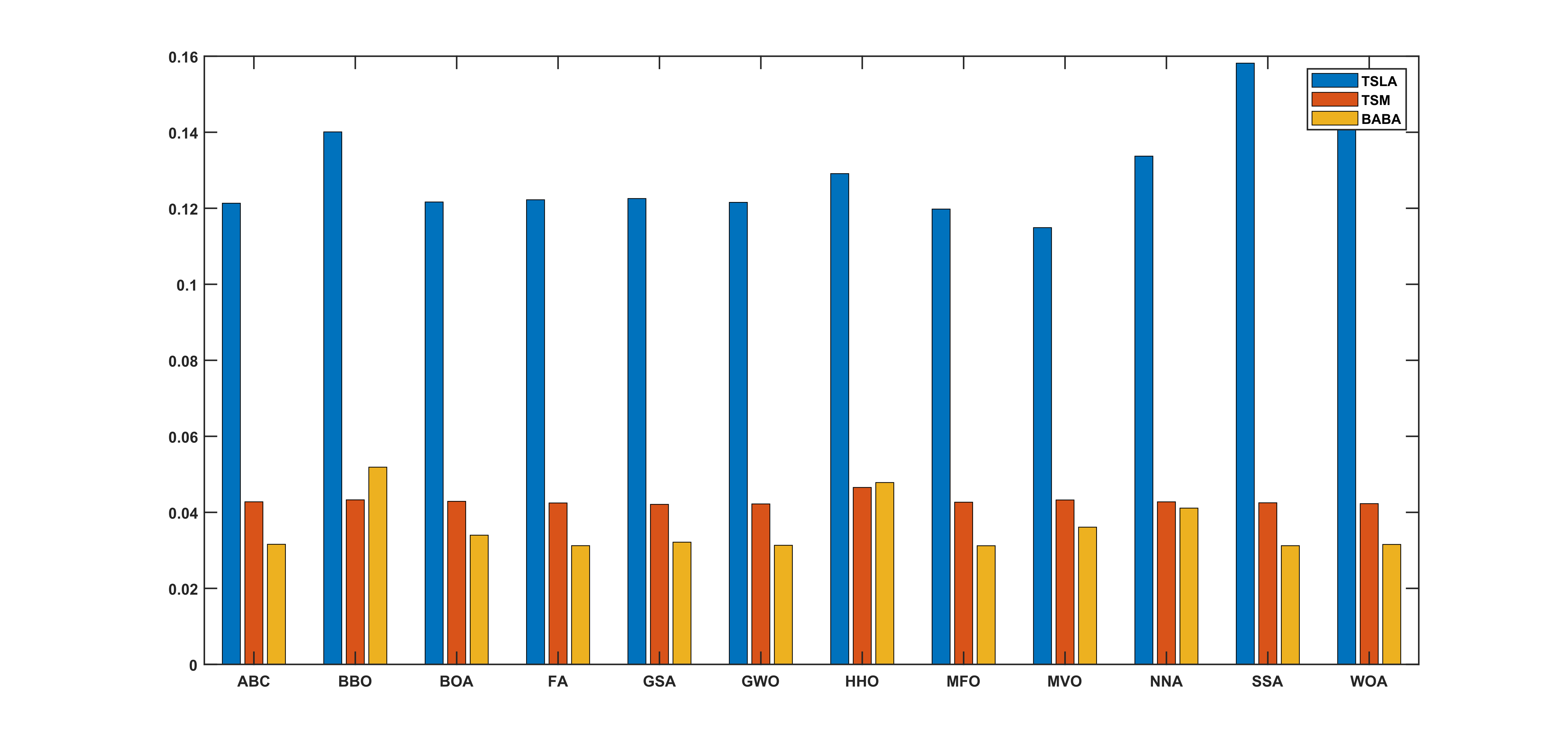}
  \caption{Bar plot of MAPEs}
 \label{mapebar}
\end{subfigure}
\newline
\centering
\begin{subfigure}{.4\textwidth}
\hspace*{-0.9in}
  \includegraphics[width=1.3\linewidth, height=4.5cm]{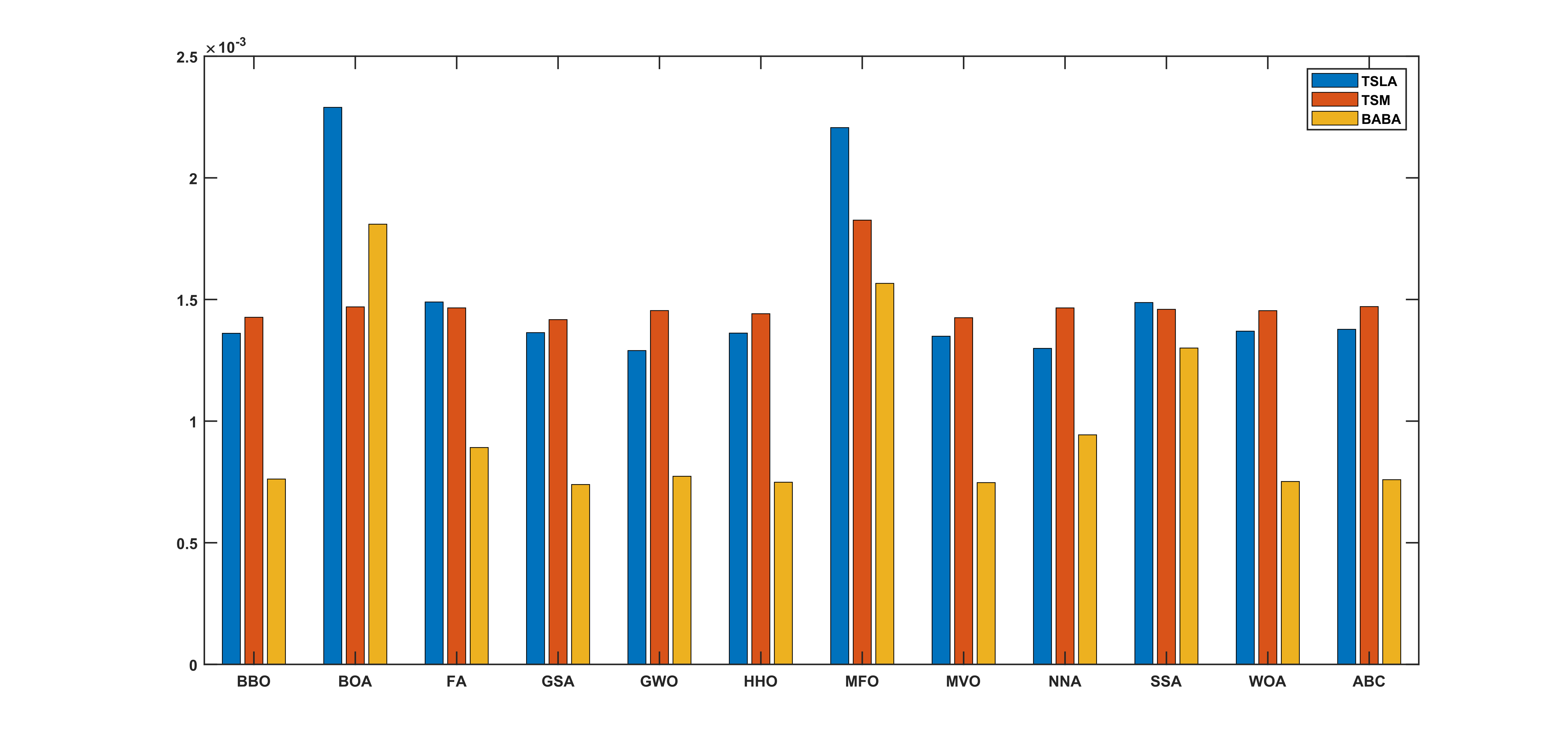}
  \centering  \caption{Bar plot of MSEs}
    \label{msebar}
\end{subfigure}
\centering
\begin{subfigure}{.4\textwidth}
\hspace*{-0.1in}
  \includegraphics[width=1.3\linewidth, height=4.5cm]{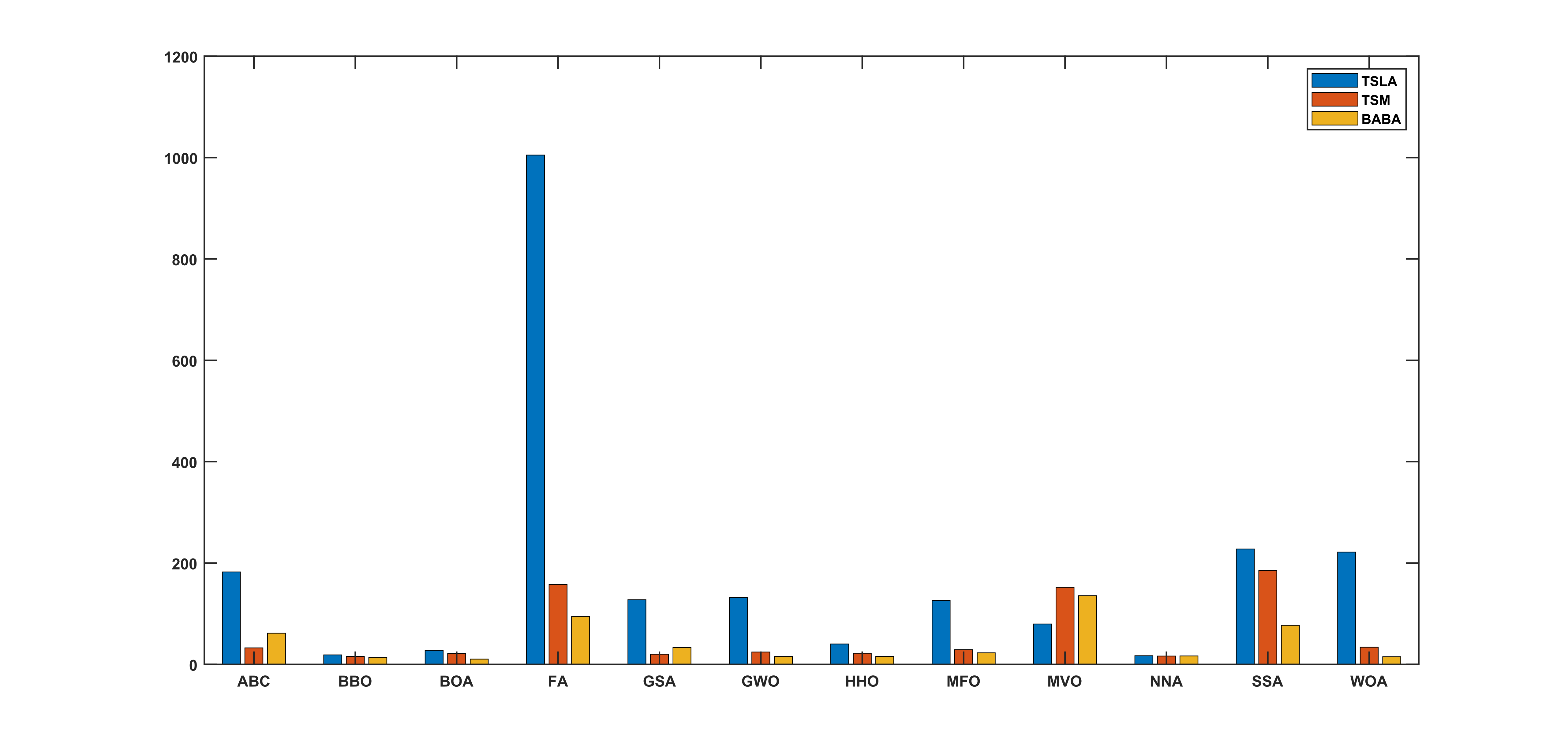}
  \centering \caption{Bar plot of Time Consumption}
     \label{timebar}
\end{subfigure}
\caption{Fig. \ref{mapebar} and Fig. \ref{msebar} compare the MAPE and the MSE bar time plots of the twelve methods. Also Fig. \ref{timebar} compare the cost time.}
\label{bartests}
\end{figure}

Also, Fig. \ref{mse} illustrates that GSA-SVR, FA-SVR, ABC-SVR, MFO-SVR, GWO-SVR and SSA-SVR achieved the best MSE accuracy below 0.0012 in comparison with all the other methods. Thus it shows that the proposed algorithm is one of the best based on MSE error. Finally, Fig. \ref{mapebar}, Fig. \ref{msebar} and Fig. \ref{timebar} respectively depict the bar plots of MAPE , MSE and cost time. Based on Fig. \ref{timecosts}, the average time consumption of the GSA-SVR model ranked seventh but competitively close to the other methods. Also it is important to mention that although BBO-SVR, BOA-SVR, HHO-SVR and NNA-SVR algorithms are computationally less expensive, but their MSE and MAPE accuracy are very bad in comparision with GSA-SVR algorithm as it is shown in Fig. \ref{mape} and Fig. \ref{mse}.

Also to compare the predictive accuracy results of our method with others here we used Diebold-Mariano test \citep{diebold2002comparing}. Base on the test, the null hypothesis of equality of any two given methods at the $5\%$ confidence level is rejected if $|DM| > 1.96$, where DM is the test statistic of the Diebold-Mariano test calculated based on the corresponding squared-error residuals.

DM-values obtained by the Diebold-Mariano test on our three stocks is presented in Table \ref{DM}. As shown in the table, there is a significant difference between our algorithm and BBO-SVR, HHO-SVR, MVO-SVR and NNA-SVR and in fact GSA-SVR is better than these algorithms and also there is no significance difference between the proposed algorithm and the rest of methods for BABA data.
For TSLA, the null hypothesis of equality is rejected for GSA-SVR and BBO-SVR, BOA-SVR, HHO-SVR and NNA-SVR and indeed, GSA-SVR based on forecasting accuracy performed better. For TSM, HHO-SVR, WOA-SVR have DM-test absolute value greater than 1.96, therefore there is only significant difference between these models and GSA-SVR. 

To summarize, based on time efficiency, MSE and MAPE measures, we conclude that our GSA-SVR algorithm is capable to find the
optimal values of the SVR parameters and can yield promising results and also is one of the best models among the same other mata-heuristics based SVR methods studied. 

\begin{table}[width=1.0\linewidth, cols=6]
  \centering
  \caption{Diebold Mariano Test}
    \begin{tabular*}{\tblwidth}{@{}CCCCCC@{} }
    \toprule
    \textbf{Methods} & \multicolumn{1}{l}{ \textbf{BABA} } & & \multicolumn{1}{l}{    \textbf{TSLA} } & & \multicolumn{1}{l}{    \textbf{TSM} } \\
    \midrule
    \textbf{ABC-SVR}   & -0.2507 & & 1.2437 & & -0.5011 \\
    \textbf{BBO-SVR}   & 5.8814 & & 5.0089 & & 0.2784 \\
    \textbf{BOA-SVR}   & 1.8033 & & 2.8169 & & 0.2922 \\
    \textbf{FA-SVR}    & -0.7461 & & 1.2767 & & -0.6720 \\
    \textbf{GWO-SVR}   & -0.5394 & & 1.2555 & & -0.6085 \\
    \textbf{HHO-SVR}   & 7.3394 & & 3.9379 & & 2.5401 \\
    \textbf{MFO-SVR}   & -0.5634 & & 1.0731 & & -0.5184 \\
    \textbf{MVO-SVR}   & 3.9999 & & 0.2153 & & 0.1945 \\
    \textbf{NNA-SVR}   & 3.8801 & & 2.5598 & & 0.1225 \\
    \textbf{WOA-SVR}   & -0.3112 & & 1.5994 & & 2.1655 \\
    \textbf{SSA-SVR}   & 1.3241 & & 1.4862 & & -0.0073 \\
    \bottomrule
    \end{tabular*}%
  \label{DM}%
\end{table}%

\clearpage
\section{Conclusion and Future Research}\label{con}

In support vector regression, parameters namely, penalty factor, $C$, RBF kernel function width parameter, $\gamma$ and radius of the epsilon tube, $\epsilon$ can change the performance of the algorithm considerably.Therefore, there is a need to optimize the parameters in a appropriate way. In this study, a novel hybrid method based on Support Vector Regression and Golden Sine algorithm is presented for selection of above mentioned parameters. The proposed method were tested on three financial time series of technology based companies, Alibaba Group Holding Limited, Tesla, Inc. and Taiwan Semiconductor Manufacturing Company Limited, using their daily closing stock market prices. For invalidation, the results are compared with eleven other meta-heuristics algorithms based SVR. According to the experimental results, GSA-SVR is capable of tuning the parameters efficiently in terms of computational time, MSE and MAPE errors.

\clearpage
\newpage

\appendix
\section{Appendix}

\begin{table}[width=1.0\linewidth,cols=4, pos=h]
  \centering
  \caption{Optimized parameters for BABA data.}
  \scalebox{0.90}{
\begin{tabular*}{\tblwidth}{@{} LLLLLLL@{} }    
    \toprule
    \textbf{Models} & \textbf{ C} & $\mathbf{\gamma}$ & $\mathbf{\epsilon}$ & \textbf{MSE } & \textbf{MAPE } & \textbf{Cost time} \\
    \midrule
    \textbf{ABC-SVR } & 113.2752 & 0.002117915 & 0.01608924 & 0.00076193 & 0.03159439 & 61.5194 \\
    \textbf{BBO-SVR } & 0.257373 & 3.12473 & 6.10E-05 & 0.00180953 & 0.0518886 & 13.957 \\
    \textbf{BOA-SVR } & 41.3041 & 0.00109727 & 0.0124801 & 0.00089146 & 0.0339914 & 10.3919 \\
    \textbf{FA-SVR } & 254.4579 & 0.001383811 & 0.008534498 & 0.0007394 & 0.0312368 & 94.71895 \\
    \textbf{GSA-SVR } & 1.25782  & 0.220345 & 0.000172123 & 0.00077329 & 0.0321525 & 33.0719 \\
    \textbf{GWO-SVR } & 254.6702 & 0.00127615 & 0.01754627 & 0.00074881 & 0.03133175 & 15.51003 \\
    \textbf{HHO-SVR} & 246.7128 & 0.9452953 & 0.06805752 & 0.00156628 & 0.04784143 & 15.86751 \\
    \textbf{MFO-SVR } & 251.2928 & 0.001859763 & 0.01025836 & 0.00074726 & 0.03121457 & 22.78994 \\
    \textbf{MVO-SVR } & 251.3184 & 0.05405353 & 0.009142361 & 0.00094357 & 0.036104 & 135.5701 \\
    \textbf{NNA-SVR} & 256   & 6.10E-05 & 0.006907721 & 0.00130049 & 0.04111579 & 16.66189 \\
    \textbf{SSA-SVR } & 254.9338 & 0.0041949 & 0.0123411 & 0.00075183 & 0.03122928 & 77.0712 \\
    \textbf{WOA-SVR } & 246.8061 & 0.001580238 & 0.01645499 & 0.00075927 & 0.0315578 & 15.11345 \\
    \bottomrule
    \end{tabular*}%
    }
  \label{babatable}%
\end{table}%
\begin{table}[width=1.0\linewidth,cols=4, pos=h]
  \centering
  \caption{Optimized parameters for TSLA data.}
  \scalebox{0.90}{
\begin{tabular*}{\tblwidth}{@{} LLLLLLL@{} }    
\toprule
    \textbf{Models} & \textbf{ C} & $\mathbf{\gamma}$ & $\mathbf{\epsilon}$ & \textbf{MSE } & \textbf{MAPE } & \textbf{Cost time} \\
    \midrule
    \textbf{ABC-SVR } & 256   & 0.0931836 & 0.01365015 & 0.00136099 & 0.1213265 & 182.4114 \\
    \textbf{BBO-SVR } & 7.85094 & 0.9454 & 6.10E-05 & 0.00228979 & 0.140091 & 18.7108 \\
    \textbf{BOA-SVR } & 0.807825 & 0.751053 & 0.00616621 & 0.00148971 & 0.121654 & 27.5524 \\
    \textbf{FA-SVR } & 254.41284 & 0.098945875 & 0.01357314 & 0.00136376 & 0.12223358 & 1004.9159 \\
    \textbf{GSA-SVR } & 26.58903 & 0.1041562 & 0.00074545 & 0.00129006 & 0.1225588 & 127.6254 \\
    \textbf{GWO-SVR } & 255.0317 & 0.0955308 & 0.01363054 & 0.00136184 & 0.1215525 & 132.128 \\
    \textbf{HHO-SVR} & 197.209 & 0.4205269 & 0.01706072 & 0.00220655 & 0.1291157 & 40.27025 \\
    \textbf{MFO-SVR } & 194.3398 & 0.08970023 & 0.01441153 & 0.00134894 & 0.1197894 & 126.3 \\
    \textbf{MVO-SVR } & 181.4413 & 0.05951673 & 0.01497712 & 0.00129902 & 0.1149064 & 79.70025 \\
    \textbf{NNA-SVR} & 253.9341 & 0.1465462 & 0.00812801 & 0.00148744 & 0.1337156 & 17.06885 \\
    \textbf{SSA-SVR } & 255.3342 & 0.1333912 & 0.00467436 & 0.00136979 & 0.1581744 & 227.6579 \\
    \textbf{WOA-SVR } & 256   & 0.1364358 & 0.00424138 & 0.00137741 & 0.1548629 & 221.5676 \\
    \bottomrule
    \end{tabular*}%
    }
  \label{tslatable}%
\end{table}%
\begin{table}[width=1.0\linewidth,cols=7, pos=h]
  \centering
  \caption{Optimized parameters for TSM data.}
  \scalebox{0.90}{
\begin{tabular*}{\tblwidth}{@{} LLLLLLL@{} }    
    \toprule
    \textbf{Models} & \textbf{ C} & $\mathbf{\gamma}$ & $\mathbf{\epsilon}$ & \textbf{MSE } & \textbf{MAPE } & \textbf{Cost time} \\
    \midrule
    \textbf{ABC-SVR } & 256   & 0.001393767 & 0.0320872 & 0.001427 & 0.042772 & 32.55109 \\
    \textbf{BBO-SVR } & 13.4353 & 0.14468 & 6.10E-05 & 0.00147 & 0.043286 & 15.6106 \\
    \textbf{BOA-SVR } & 51.0281 & 0.0233701 & 0.0106542 & 0.001466 & 0.042897 & 21.3761 \\
    \textbf{FA-SVR } & 145.3738 & 0.004001169 & 0.03353944 & 0.001417 & 0.042468 & 157.6236 \\
    \textbf{GSA-SVR } & 1.4889 & 0.0970193 & 0.00213057 & 0.001454 & 0.042101 & 20.1198 \\
    \textbf{GWO-SVR } & 6.35558 & 0.0320962 & 0.00284335 & 0.001442 & 0.042212 & 24.3695 \\
    \textbf{HHO-SVR} & 171.9925 & 0.9798621 & 0.1031784 & 0.001826 & 0.04655 & 22.02656 \\
    \textbf{MFO-SVR } & 230.3153 & 0.003261379 & 0.04130702 & 0.001425 & 0.042671 & 28.87215 \\
    \textbf{MVO-SVR } & 252.9431 & 0.03697545 & 6.10E-05 & 0.001465 & 0.043254 & 151.9899 \\
    \textbf{NNA-SVR} & 256   & 0.01576342 & 0.0005394 & 0.00146 & 0.042764 & 16.40153 \\
    \textbf{SSA-SVR } & 255.9636 & 0.004876394 & 0.00013112 & 0.001454 & 0.042515 & 185.3799 \\
    \textbf{WOA-SVR } & 1.76772 & 0.0728386 & 6.10E-05 & 0.001471 & 0.04229 & 33.9208 \\
    \bottomrule
    \end{tabular*}%
    }
  \label{tsmtable}%
\end{table}%

\newpage
\bibliographystyle{model2-names}
\bibliography{refs}

\begin{thebibliography}{31}
\expandafter\ifx\csname natexlab\endcsname\relax\def\natexlab#1{#1}\fi
\providecommand{\url}[1]{\texttt{#1}}
\providecommand{\href}[2]{#2}
\providecommand{\path}[1]{#1}
\providecommand{\DOIprefix}{doi:}
\providecommand{\ArXivprefix}{arXiv:}
\providecommand{\URLprefix}{URL: }
\providecommand{\Pubmedprefix}{pmid:}
\providecommand{\doi}[1]{\href{http://dx.doi.org/#1}{\path{#1}}}
\providecommand{\Pubmed}[1]{\href{pmid:#1}{\path{#1}}}
\providecommand{\bibinfo}[2]{#2}
\ifx\xfnm\relax \def\xfnm[#1]{\unskip,\space#1}\fi
\bibitem[{Abarbanel(2012)}]{abarbanel2012analysis}
\bibinfo{author}{Abarbanel, H.}, \bibinfo{year}{2012}.
\newblock \bibinfo{title}{Analysis of observed chaotic data}.
\newblock \bibinfo{publisher}{Springer Science \& Business Media}.
\bibitem[{Blum and Roli(2003)}]{blum2003metaheuristics}
\bibinfo{author}{Blum, C.}, \bibinfo{author}{Roli, A.}, \bibinfo{year}{2003}.
\newblock \bibinfo{title}{Metaheuristics in combinatorial optimization:
  Overview and conceptual comparison}.
\newblock \bibinfo{journal}{ACM computing surveys (CSUR)} \bibinfo{volume}{35},
  \bibinfo{pages}{268--308}.
\bibitem[{Chang and Lin(2011)}]{chang2011libsvm}
\bibinfo{author}{Chang, C.C.}, \bibinfo{author}{Lin, C.J.},
  \bibinfo{year}{2011}.
\newblock \bibinfo{title}{Libsvm: A library for support vector machines}.
\newblock \bibinfo{journal}{ACM transactions on intelligent systems and
  technology (TIST)} \bibinfo{volume}{2}, \bibinfo{pages}{27}.
\bibitem[{Chapelle et~al.(2002)Chapelle, Vapnik, Bousquet and
  Mukherjee}]{chapelle2002choosing}
\bibinfo{author}{Chapelle, O.}, \bibinfo{author}{Vapnik, V.},
  \bibinfo{author}{Bousquet, O.}, \bibinfo{author}{Mukherjee, S.},
  \bibinfo{year}{2002}.
\newblock \bibinfo{title}{Choosing multiple parameters for support vector
  machines}.
\newblock \bibinfo{journal}{Machine learning} \bibinfo{volume}{46},
  \bibinfo{pages}{131--159}.
\bibitem[{Chen and Yang(2012)}]{chen2012multiscale}
\bibinfo{author}{Chen, Y.}, \bibinfo{author}{Yang, H.}, \bibinfo{year}{2012}.
\newblock \bibinfo{title}{Multiscale recurrence analysis of long-term nonlinear
  and nonstationary time series}.
\newblock \bibinfo{journal}{Chaos, Solitons \& Fractals} \bibinfo{volume}{45},
  \bibinfo{pages}{978--987}.
\bibitem[{Cortes and Vapnik(1995)}]{cortes1995support}
\bibinfo{author}{Cortes, C.}, \bibinfo{author}{Vapnik, V.},
  \bibinfo{year}{1995}.
\newblock \bibinfo{title}{Support-vector networks}.
\newblock \bibinfo{journal}{Machine learning} \bibinfo{volume}{20},
  \bibinfo{pages}{273--297}.
\bibitem[{Diebold and Mariano(2002)}]{diebold2002comparing}
\bibinfo{author}{Diebold, F.X.}, \bibinfo{author}{Mariano, R.S.},
  \bibinfo{year}{2002}.
\newblock \bibinfo{title}{Comparing predictive accuracy}.
\newblock \bibinfo{journal}{Journal of Business \& economic statistics}
  \bibinfo{volume}{20}, \bibinfo{pages}{134--144}.
\bibitem[{Duan et~al.(2003)Duan, Keerthi and Poo}]{duan2003evaluation}
\bibinfo{author}{Duan, K.}, \bibinfo{author}{Keerthi, S.S.},
  \bibinfo{author}{Poo, A.N.}, \bibinfo{year}{2003}.
\newblock \bibinfo{title}{Evaluation of simple performance measures for tuning
  svm hyperparameters}.
\newblock \bibinfo{journal}{Neurocomputing} \bibinfo{volume}{51},
  \bibinfo{pages}{41--59}.
\bibitem[{Ghanbari and Arian(2019)}]{ghanbari2019forecasting}
\bibinfo{author}{Ghanbari, M.}, \bibinfo{author}{Arian, H.},
  \bibinfo{year}{2019}.
\newblock \bibinfo{title}{Forecasting stock market with support vector
  regression and butterfly optimization algorithm}.
\newblock \bibinfo{journal}{arXiv preprint arXiv:1905.11462} .
\bibitem[{Gogna and Tayal(2013)}]{gogna2013metaheuristics}
\bibinfo{author}{Gogna, A.}, \bibinfo{author}{Tayal, A.}, \bibinfo{year}{2013}.
\newblock \bibinfo{title}{Metaheuristics: review and application}.
\newblock \bibinfo{journal}{Journal of Experimental \& Theoretical Artificial
  Intelligence} \bibinfo{volume}{25}, \bibinfo{pages}{503--526}.
\bibitem[{Gu et~al.(2011)Gu, Zhu and Jiang}]{gu2011housing}
\bibinfo{author}{Gu, J.}, \bibinfo{author}{Zhu, M.}, \bibinfo{author}{Jiang,
  L.}, \bibinfo{year}{2011}.
\newblock \bibinfo{title}{Housing price forecasting based on genetic algorithm
  and support vector machine}.
\newblock \bibinfo{journal}{Expert Systems with Applications}
  \bibinfo{volume}{38}, \bibinfo{pages}{3383--3386}.
\bibitem[{Hsu et~al.(2003)Hsu, Chang, Lin et~al.}]{hsu2003practical}
\bibinfo{author}{Hsu, C.W.}, \bibinfo{author}{Chang, C.C.},
  \bibinfo{author}{Lin, C.J.}, et~al., \bibinfo{year}{2003}.
\newblock \bibinfo{title}{A practical guide to support vector classification} .
\bibitem[{Huang(2012)}]{huang2012hybrid}
\bibinfo{author}{Huang, C.F.}, \bibinfo{year}{2012}.
\newblock \bibinfo{title}{A hybrid stock selection model using genetic
  algorithms and support vector regression}.
\newblock \bibinfo{journal}{Applied Soft Computing} \bibinfo{volume}{12},
  \bibinfo{pages}{807--818}.
\bibitem[{Kavousi-Fard et~al.(2014)Kavousi-Fard, Samet and
  Marzbani}]{kavousi2014new}
\bibinfo{author}{Kavousi-Fard, A.}, \bibinfo{author}{Samet, H.},
  \bibinfo{author}{Marzbani, F.}, \bibinfo{year}{2014}.
\newblock \bibinfo{title}{A new hybrid modified firefly algorithm and support
  vector regression model for accurate short term load forecasting}.
\newblock \bibinfo{journal}{Expert systems with applications}
  \bibinfo{volume}{41}, \bibinfo{pages}{6047--6056}.
\bibitem[{Keerthi et~al.(2007)Keerthi, Sindhwani and
  Chapelle}]{keerthi2007efficient}
\bibinfo{author}{Keerthi, S.S.}, \bibinfo{author}{Sindhwani, V.},
  \bibinfo{author}{Chapelle, O.}, \bibinfo{year}{2007}.
\newblock \bibinfo{title}{An efficient method for gradient-based adaptation of
  hyperparameters in svm models}, in: \bibinfo{booktitle}{Advances in neural
  information processing systems}, pp. \bibinfo{pages}{673--680}.
\bibitem[{Kennel et~al.(1992)Kennel, Brown and
  Abarbanel}]{kennel1992determining}
\bibinfo{author}{Kennel, M.B.}, \bibinfo{author}{Brown, R.},
  \bibinfo{author}{Abarbanel, H.D.}, \bibinfo{year}{1992}.
\newblock \bibinfo{title}{Determining embedding dimension for phase-space
  reconstruction using a geometrical construction}.
\newblock \bibinfo{journal}{Physical review A} \bibinfo{volume}{45},
  \bibinfo{pages}{3403}.
\bibitem[{Kwok(2000)}]{kwok2000evidence}
\bibinfo{author}{Kwok, J.T.Y.}, \bibinfo{year}{2000}.
\newblock \bibinfo{title}{The evidence framework applied to support vector
  machines}.
\newblock \bibinfo{journal}{IEEE Transactions on Neural Networks}
  \bibinfo{volume}{11}, \bibinfo{pages}{1162--1173}.
\bibitem[{Li et~al.(2018)Li, Fang and Liu}]{li2018parameter}
\bibinfo{author}{Li, S.}, \bibinfo{author}{Fang, H.}, \bibinfo{author}{Liu,
  X.}, \bibinfo{year}{2018}.
\newblock \bibinfo{title}{Parameter optimization of support vector regression
  based on sine cosine algorithm}.
\newblock \bibinfo{journal}{Expert systems with Applications}
  \bibinfo{volume}{91}, \bibinfo{pages}{63--77}.
\bibitem[{Li-Xia et~al.(2011)Li-Xia, Yi-Qi and Liu}]{li2011tax}
\bibinfo{author}{Li-Xia, L.}, \bibinfo{author}{Yi-Qi, Z.},
  \bibinfo{author}{Liu, X.y.}, \bibinfo{year}{2011}.
\newblock \bibinfo{title}{Tax forecasting theory and model based on svm
  optimized by pso}.
\newblock \bibinfo{journal}{Expert Systems with Applications}
  \bibinfo{volume}{38}, \bibinfo{pages}{116--120}.
\bibitem[{Min et~al.(2006)Min, Lee and Han}]{min2006hybrid}
\bibinfo{author}{Min, S.H.}, \bibinfo{author}{Lee, J.}, \bibinfo{author}{Han,
  I.}, \bibinfo{year}{2006}.
\newblock \bibinfo{title}{Hybrid genetic algorithms and support vector machines
  for bankruptcy prediction}.
\newblock \bibinfo{journal}{Expert systems with applications}
  \bibinfo{volume}{31}, \bibinfo{pages}{652--660}.
\bibitem[{Mustaffa et~al.(2015)Mustaffa, Sulaiman and Kahar}]{mustaffa2015ls}
\bibinfo{author}{Mustaffa, Z.}, \bibinfo{author}{Sulaiman, M.H.},
  \bibinfo{author}{Kahar, M.N.M.}, \bibinfo{year}{2015}.
\newblock \bibinfo{title}{Ls-svm hyper-parameters optimization based on gwo
  algorithm for time series forecasting}, in: \bibinfo{booktitle}{2015 4th
  International Conference on Software Engineering and Computer Systems
  (ICSECS)}, \bibinfo{organization}{IEEE}. pp. \bibinfo{pages}{183--188}.
\bibitem[{Smola and Sch{\"o}lkopf(2004)}]{smola2004tutorial}
\bibinfo{author}{Smola, A.J.}, \bibinfo{author}{Sch{\"o}lkopf, B.},
  \bibinfo{year}{2004}.
\newblock \bibinfo{title}{A tutorial on support vector regression}.
\newblock \bibinfo{journal}{Statistics and computing} \bibinfo{volume}{14},
  \bibinfo{pages}{199--222}.
\bibitem[{Takens(1981)}]{takens1981detecting}
\bibinfo{author}{Takens, F.}, \bibinfo{year}{1981}.
\newblock \bibinfo{title}{Detecting strange attractors in turbulence}, in:
  \bibinfo{booktitle}{Dynamical systems and turbulence, Warwick 1980}.
  \bibinfo{publisher}{Springer}, pp. \bibinfo{pages}{366--381}.
\bibitem[{Talbi(2009)}]{talbi2009metaheuristics}
\bibinfo{author}{Talbi, E.G.}, \bibinfo{year}{2009}.
\newblock \bibinfo{title}{Metaheuristics: from design to implementation}.
  volume~\bibinfo{volume}{74}.
\newblock \bibinfo{publisher}{John Wiley \& Sons}.
\bibitem[{Tanyildizi and Demir(2017)}]{tanyildizi2017golden}
\bibinfo{author}{Tanyildizi, E.}, \bibinfo{author}{Demir, G.},
  \bibinfo{year}{2017}.
\newblock \bibinfo{title}{Golden sine algorithm: a novel math-inspired
  algorithm}.
\newblock \bibinfo{journal}{Advances in Electrical and Computer Engineering}
  \bibinfo{volume}{17}, \bibinfo{pages}{71--79}.
\bibitem[{Tavakkoli et~al.(2015)Tavakkoli, Rezaeenour and
  Hadavandi}]{tavakkoli2015novel}
\bibinfo{author}{Tavakkoli, A.}, \bibinfo{author}{Rezaeenour, J.},
  \bibinfo{author}{Hadavandi, E.}, \bibinfo{year}{2015}.
\newblock \bibinfo{title}{A novel forecasting model based on support vector
  regression and bat meta-heuristic (bat--svr): case study in printed circuit
  board industry}.
\newblock \bibinfo{journal}{International Journal of Information Technology \&
  Decision Making} \bibinfo{volume}{14}, \bibinfo{pages}{195--215}.
\bibitem[{Thissen et~al.(2003)Thissen, Van~Brakel, De~Weijer, Melssen and
  Buydens}]{thissen2003using}
\bibinfo{author}{Thissen, U.}, \bibinfo{author}{Van~Brakel, R.},
  \bibinfo{author}{De~Weijer, A.}, \bibinfo{author}{Melssen, W.},
  \bibinfo{author}{Buydens, L.}, \bibinfo{year}{2003}.
\newblock \bibinfo{title}{Using support vector machines for time series
  prediction}.
\newblock \bibinfo{journal}{Chemometrics and intelligent laboratory systems}
  \bibinfo{volume}{69}, \bibinfo{pages}{35--49}.
\bibitem[{Vapnik(2013)}]{vapnik2013nature}
\bibinfo{author}{Vapnik, V.}, \bibinfo{year}{2013}.
\newblock \bibinfo{title}{The nature of statistical learning theory}.
\newblock \bibinfo{publisher}{Springer science \& business media}.
\bibitem[{Wu et~al.(2009)Wu, Tzeng and Lin}]{wu2009novel}
\bibinfo{author}{Wu, C.H.}, \bibinfo{author}{Tzeng, G.H.},
  \bibinfo{author}{Lin, R.H.}, \bibinfo{year}{2009}.
\newblock \bibinfo{title}{A novel hybrid genetic algorithm for kernel function
  and parameter optimization in support vector regression}.
\newblock \bibinfo{journal}{Expert Systems with Applications}
  \bibinfo{volume}{36}, \bibinfo{pages}{4725--4735}.
\bibitem[{Wu(2010)}]{wu2010hybrid}
\bibinfo{author}{Wu, Q.}, \bibinfo{year}{2010}.
\newblock \bibinfo{title}{A hybrid-forecasting model based on gaussian support
  vector machine and chaotic particle swarm optimization}.
\newblock \bibinfo{journal}{Expert Systems with Applications}
  \bibinfo{volume}{37}, \bibinfo{pages}{2388--2394}.
\bibitem[{Yeh et~al.(2011)Yeh, Huang and Lee}]{yeh2011multiple}
\bibinfo{author}{Yeh, C.Y.}, \bibinfo{author}{Huang, C.W.},
  \bibinfo{author}{Lee, S.J.}, \bibinfo{year}{2011}.
\newblock \bibinfo{title}{A multiple-kernel support vector regression approach
  for stock market price forecasting}.
\newblock \bibinfo{journal}{Expert Systems with Applications}
  \bibinfo{volume}{38}, \bibinfo{pages}{2177--2186}.

\end{thebibliography}

\end{document}